\documentclass{article} 
\usepackage{iclr2024_conference,times}

\usepackage{hyperref}
\usepackage{url}
\usepackage{fancyhdr}

\fancypagestyle{firstpage}{
    \fancyhf{} 
    \fancyhead[L]{Preprint}
    \fancyfoot[L]{\textcopyright \ 2023 Christian A. Schiller.}

}

\usepackage{algorithm}
\usepackage{algorithmic}
\usepackage{amsthm}
\usepackage{caption}
\usepackage{subcaption}
\usepackage{mathtools}
\usepackage{adjustbox}
\usepackage{multirow}
\usepackage{siunitx}

\usepackage{enumitem}
\setlist{leftmargin=3mm}

\usepackage{wrapfig,lipsum,booktabs}
\usepackage{needspace}

\title{Virtual Augmented Reality \\ for Atari Reinforcement Learning}

\author{%
    Christian A. Schiller \\
    University of Duisburg-Essen, 45141 Essen, Germany \\
    christian.schiller@stud.uni-due.de \\
}

\iclrfinalcopy 
\begin{document}

\thispagestyle{firstpage} 

\maketitle

\begin{abstract}
Reinforcement Learning (RL) has achieved significant milestones in
the gaming domain, most notably Google DeepMind's AlphaGo defeating
human Go champion Ken Jie. This victory was also made possible through
the Atari Learning Environment (ALE): The ALE has been foundational
in RL research, facilitating significant RL algorithm developments
such as AlphaGo and others. In current Atari video game RL research,
RL agents' perceptions of its environment is based on raw pixel data
from the Atari video game screen with minimal image preprocessing.
Contrarily, cutting-edge ML research, external to the Atari video
game RL research domain, is focusing on enhancing image perception.
A notable example is Meta Research's ``Segment Anything Model''
(SAM), a foundation model capable of segmenting images without prior
training (zero-shot). This paper addresses a novel methodical question:
Can state-of-the-art image segmentation models such as SAM improve
the performance of RL agents playing Atari video games? The results
suggest that SAM can serve as a ``virtual augmented reality'' for
the RL agent, boosting its Atari video game playing performance under
certain conditions. Comparing RL agent performance results from raw
and augmented pixel inputs provides insight into these conditions.
Although this paper was limited by computational constraints, the
findings show improved RL agent performance for augmented pixel inputs
and can inform broader research agendas in the domain of ``virtual
augmented reality for video game playing RL agents''.
\end{abstract}

\section{Introduction}

Reinforcement Learning (RL) is the third major discipline of machine
learning alongside supervised and unsupervised learning. RL follows
a ``trial and error'' principle, similar to human learning from
making physical experiences rather than from reading books. In RL,
an agent tries different actions in a simulated (virtual) environment.
Evaluating the agent's actions against a specified goal, the agent
is rewarded for successes and punished for failures. Virtual video
game environments are widely used for researching RL algorithms, because
they can be simulated in faster than real time. This speeds up the
collection of training data (experiences in the environment). For
example, in the Atari video game ``Breakout'', the player's goal
is to dismantle a wall on the top of the screen brick by brick with
a moving ball, using a racket which can only move left or right at
the bottom of the screen. Using RL to master this game means that
the agent takes control of the racket in place of the human player
and tries different actions (move racket left or right). The agent
is rewarded when destroying bricks and punished when missing the ball.
The RL algorithm controlling the agent's behavior tries this until
the agent has learned to move the racket so well as to destroy all
bricks and never miss the ball. Depending on the RL algorithm, tens
of thousands or even millions of tries of playing the game are required
to achieve at least the same result as a human. A human knows from
touching a hot stove just once that it is not a recommended action
to try again. A human can intuitively grasp an easy and simplistic
game such as ``Breakout'' and will achieve high scores already after
two or three tries instead of thousands or millions or tries.

Many prominent RL successes, such as Google DeepMind's AlphaGo RL
algorithm defeat of the human Go champion Ken Jie (\cite{silver2016}),
are based on ALE, the Atari Learning Environment (\cite{bellemare2013}).
The ALE allowed RL algorithms to assume the player role in Atari VCS
2600 video games easily by wrapping an Atari VCS 2600 emulator into
an API well suited for supporting the RL workflow. AlphaGo's RL algorithm
development was refined using the ALE (\cite{mnih2013}). The current
state-of-the-art Go-Explore RL algorithm was even able to beat the
human world record for the difficult Atari VCS 2600 game ``Montezuma\textquoteleft s
Revenge'' (\cite{ecoffet2021}). Therefore, it can be concluded that
the research efforts in the domain ``RL in video games'' are foundational
for RL algorithm research. A recent analysis on the evolution of RL
algorithms for video games shows that the RL algorithm itself has
a very large impact on performance (\cite{ecoffet2021}). RL algorithms
became steadily better over time. More recent algorithms like Agent57
or Go-Explore from 2020 beat the Atari game scores achieved by the
2015 DQN algorithm DQN by factors of 10 or more (\cite{ecoffet2021}). 

While the RL algorithms evolved, the environment state used as input
to the RL algorithms stayed the same: The raw pixel input of the video
game environment as displayed by the video game platform (\cite{justesen2020}).
The perception of the RL agent used in RL video game research is limited
to raw image data without additional semantic information. Even the
current state of the art video game RL pipeline only uses a simple
two-step preprocessing on the raw image: Downscale image from 210x160
to 84x84 pixels, and grayscale image (\cite{machado2018}), before
being further processed by a five-layer convolutional neural network.

Current state-of-the-art ML research from other domains than video
RL addresses this perception problem by developing ML models which
can add a first layer of semantics to a raw pixel image by performing
a segmentation of the image into distinct areas or objects. A prominent
example for this new class of ML models is Meta Research's Segment
Anything Model (SAM), which performs image segmentation (\cite{kirillov2023}).
The SAM model is considered a foundational ML model and can be used
even without additional ML training processes (zero shot usage). The
findings that currently ``RL for video games'' is always based on
raw pixel input of the video game platform, plus the availability
of new state-of-the ML research like SAM, leads to the research question
in scope of this paper:

\emph{}%
\noindent\fbox{\begin{minipage}[t]{1\columnwidth - 2\fboxsep - 2\fboxrule}%
\emph{How does the use of state-of-the art image segmentation ML models
like SAM on video game raw pixel inputs impact the performance of
an RL agent for playing video games?}%
\end{minipage}}

The hypothesis for this research is that an improvement of game playing
performance of the RL agent can be observed, when the agent's perception
(the raw pixel input) is augmented with image segmentation. Such augmented
pixel input constitutes a ``virtual augmented reality'' for the
RL agent. The RL agent is trained twice. Once without image segmentation
(only with raw pixels), and once with image segmentation. Game playing
performance is compared. The observed performance could differ from
one RL algorithm to another or from one video game to another. Due
to the limited computing infrastructure availability and time+effort
scope of this research, only a limited set of Atari VCS 2600 games,
one RL algorithm and one image segmentation algorithm could be evaluated.
The initial results of this research could be used to scale the research
question to more RL algorithms, more image segmentation algorithms,
more video games and more video game platforms.

\section{Research Methodology}

To explore the research question, an RL pipeline for Atari video games
is implemented, following the guidance for training and evaluating
RL agents for Atari games from the paper ``Revisiting the Atari Learning
Environment'' (\cite{machado2018}). The research question was triggered
by the release of Meta Research's SAM (``Segment Anything Model'')
image segmentation model, which allows zero shot image segmentation.
Therefore the SAM model is integrated into the RL pipeline. The RL
pipeline is applied to a limited set of Atari games, with and without
SAM-based image segmentation. The Atari games are chosen based on
an existing taxonomy for Atari games (\cite{bellemare2016}). An additional
criterion which is assumed to be relevant for the research question
is introduced: The number of objects a game typically displays on
screen. The number of chosen games for experiments is matched to the
capabilities of the available infrastructure (i.e. available computing
resources and time for computing). To achieve comparability of training
runs, the RL training hyperparameters are aligned, i.e. are set to
the same values for both runs (with and without image preprocessing).
The only difference is the image segmentation step. This paper's research
focus is visualized in the following figure, depicting the typical
RL video game training pipeline for Atari games (\cite{machado2018}).
\begin{center}
\begin{figure}[H]
\begin{centering}
\includegraphics[width=14cm]{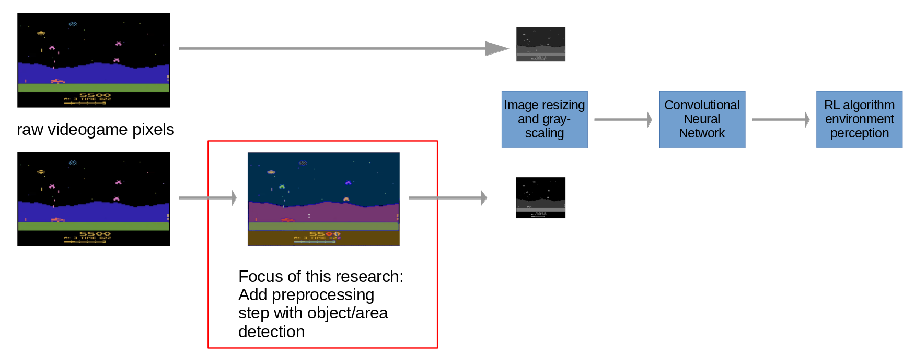}
\par\end{centering}
\caption{Focus of this research visualized in a typical RL video game training
pipeline}
\end{figure}
\par\end{center}

\section{Implementation}

For the RL pipeline baseline, a combined installation of CleanRL (\cite{huang2022}),
Gymnasium (Farama Foundation 2023\footnote{https://farama.org/})
and the Atari Learning Environment (\cite{machado2018}) are used.
The rationale for this choice is that these components are already
pre-integrated within the CleanRL codebase and work out of the box,
whereas combinations of other components would have required integration
effort seen outside the scope of the research question. CleanRL provides
an RL training pipeline template for Atari games, and the algorithm
implementation is performance optimized for the available computing
infrastructure for the project. Gymnasium has a wrapper architecture
which allows to add arbitrary processing of observations before they
are processed by the RL algorithm. As RL algorithm, PPO (Proximal
Policy Optimization) is chosen, due to its known good general performance
for Atari video game environments (\cite{justesen2020}). The references
for the chosen RL baseline and the implementation for this paper are
given in the appendix.

The SAM model is published in three versions which differ in model
sizes: ``vit\_h'' (largest), ``vit\_l'' (medium), ``vit\_b''
(smallest). Due to the limited GPU infrastructure available for the
research project (2021 NVIDIA Quadro RTX 3000 GPU with 6GB Memory),
only the smallest model version ``vit\_b'' could be used. To verify
SAM ``vit\_b'' model preprocessing integration into the RL pipeline,
a function for integrating SAM into the image processing pipeline
is implemented which uses the SAM model on every observation the RL
agent makes. It is verified that the SAM model usage is working by
saving an image file from the running pipeline of training the Atari
2600 game ``Breakout'' with the PPO algorithm implementation from
CleanRL.

Execution of test case:\\\\
\texttt{ppo\_atari\_sam.py --env-id='Breakout-v5' --clip-coef=0.2 --num-envs=1 --num-minibatches=8 --num-steps=128 --update-epochs=3 --capture-video}

\newpage{}

\begin{figure}
\begin{centering}
\includegraphics[width=10cm]{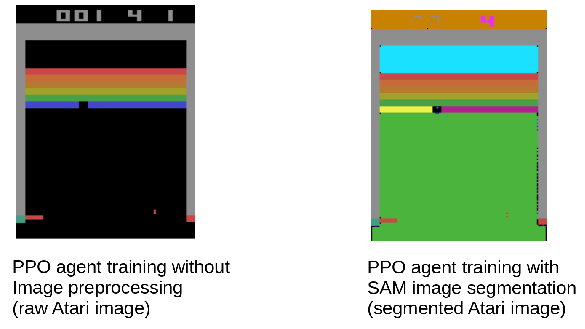}
\par\end{centering}
\caption{SAM image segmentation model integrated in RL pipeline - left: raw
Atari image, right: SAM model inference result (i.e. resulting image
segmentation)}
\end{figure}

The figure shows the implementation of the SAM image segmentation
model for the Atari game ``Breakout''. Due to the nature of the
game screen -- quite static, with only the ball and the racket as
moving objects -- apart from the color changes, the image segments
are not much different than the original image. 

Introducing the SAM model inference (\textasciitilde 0.7sec) into
the RL pipeline reduces the performance of the RL algorithm in terms
of ``game scenes processed per second'' severely. In order to counterweight
this computing time performance loss, the method implementation has
to be adapted for computational efficiency to match the available
GPU infrastructure. For evaluating one game with SAM model, no more
than 24 hours of computation should be spent. In order to achieve
this, several adjustments have to be made, resulting from CleanRL
and SAM hyperparameter tuning experimentation. The Atari frameskip
is set to 4, i.e. only 15 instead of 60 frames per second have to
processed, at the risk of missing game events happening in the missing
frames. The PPO algorithm learning rate is increased to 0.0025. The
PPO algorithm update policy (epsilon) is slightly increased to 0.25.
The agent training time is capped at 20.000 global steps for both
training runs; this corresponds to about 24 hours of runtime for the
SAM model training run on the available GPU infrastructure, and only
minutes of training without SAM.

Execution of RL training for each game without SAM image segmentation:\\\\
\texttt{ppo\_atari.py -seed=47 --env-id='ALE/<Game>-v5' --clip-coef=0.25 --learning-rate=2.5e-3 --num-envs=1 --num-minibatches=8 --num-steps=128 --update-epochs=2 --total-timesteps=20000 --capture-video}

Execution of RL training for each game with SAM image segmentation:\\\\
\texttt{ppo\_atari\_sam.py -seed=47 --env-id='ALE/<Game>-v5' --clip-coef=0.25 --learning-rate=2.5e-3 --num-envs=1 --num-minibatches=8 --num-steps=128 --update-epochs=2 --total-timesteps=20000 --capture-video}

Atari 2600 games can be categorized into four distinct categories
(\cite{bellemare2016}). The finding from integrating SAM into the
RL pipeline -- for some games, the SAM segmentation is not much different
than the original image -- leads to adding an additional criterion
for selecting games for experiments. For the research question, an
additional discriminator is added: Number of objects typically visible
on screen (low or high), in order to evaluate if that is an additional
factor in performance of resulting RL agent, as found when testing
the game ``Breakout''.

For the research question in scope, we test two games from each category,
with the exception of the ``hard exploration -- sparse reward''
category. The rationale for excluding this category is that the limited
number of timesteps available for training -- only 20.000 -- is
in general not sufficient to learn an RL agent for this type of games;
typically tens to hundreds of millions of game frames have to be processed
for this game category (\cite{machado2018}).

The following table outlines the full scope of the experiments performed.
For each of the listed twelve Atari games, two RL agents are trained,
one without SAM and one with SAM. Afterwards, the RL agent gaming
performance is evaluated.

\begin{table}[H]
\begin{tabular}{|c|c|c|c|c|}
\hline 
\textbf{\small{}Bellemare taxonomy \textrightarrow{}} & \multicolumn{2}{c|}{\textbf{\small{}Easy exploration}} & \multicolumn{2}{c|}{\textbf{\small{}Hard exploration}}\tabularnewline
\cline{2-5} \cline{3-5} \cline{4-5} \cline{5-5} 
\textbf{\small{}New discriminator \textdownarrow{}} & \textbf{\small{}Human-optimal} & \textbf{\small{}Score exploit} & \textbf{\small{}Dense reward} & \textbf{\small{}Sparse reward}\tabularnewline
\hline 
\textbf{\small{}Low number of} & {\small{}Breakout} & {\small{}Kung Fu Master} & {\small{}Ms. Pac-Man} & {\small{}-}\tabularnewline
\textbf{\small{}on-screen objects} & {\small{}Pong} & {\small{}Road Runner} & {\small{}Q{*}Bert} & {\small{}-}\tabularnewline
\hline 
\textbf{\small{}High number of} & {\small{}Space Invaders} & {\small{}Seaquest} & {\small{}Frostbite} & {\small{}-}\tabularnewline
\textbf{\small{}on-screen objects} & {\small{}Chopper Command} & {\small{}Beam Rider} & {\small{}Zaxxon} & {\small{}-}\tabularnewline
\hline 
\end{tabular}

\caption{Atari 2600 games chosen for experiment}
\end{table}

\section{Results}

The result table lists the RL agents' game scores achieved after 20.000
global steps. Detailed results and discussion of SAM performance on
the individual games are listed in full detail in the appendix. For
each training run, the training times are recorded as well, which
show the computing time impact of SAM. For each game, example frames
of raw Atari image and SAM-processed Atari image are provided. The
following table summarizes the relevant score results from the experiments,
ordered by percentage of game score improvement from SAM-augmented
agent over raw pixel agent.

\begin{table}[H]
\begin{tabular}{|c|c|c|c|}
\hline 
\multirow{3}{*}{\textbf{\small{}Game}} & \textbf{\scriptsize{}Game score improvement} & \textbf{\scriptsize{}SAM-augmented agent} & \textbf{\scriptsize{}Raw pixel agent}\tabularnewline
 & \textbf{\scriptsize{}of SAM-augmented agent} & \textbf{\scriptsize{}game score achived} & \textbf{\scriptsize{}game score achieved}\tabularnewline
 & \textbf{\scriptsize{}over raw pixel agent} &  & \tabularnewline
\hline 
\hline 
\textbf{\small{}Beam Rider} & \textbf{\small{}129.4\%} & {\small{}505.1} & {\small{}390.3}\tabularnewline
\hline 
\textbf{\small{}Seaquest} & \textbf{\small{}105.6\%} & {\small{}230.4} & {\small{}218.1}\tabularnewline
\hline 
\textbf{\small{}Chopper Command} & \textbf{\small{}104.8\%} & {\small{}932.2} & {\small{}889.6}\tabularnewline
\hline 
\textbf{\small{}Space Invaders} & \textbf{\small{}101.3\%} & {\small{}226.2} & {\small{}223.3}\tabularnewline
\hline 
\textbf{\small{}Kung Fu Master} & {\small{}93.4\%} & {\small{}73.2} & {\small{}78.4}\tabularnewline
\hline 
\textbf{\small{}Q{*}Bert} & {\small{}82.3\%} & {\small{}375.8} & {\small{}456.6}\tabularnewline
\hline 
\textbf{\small{}Ms. Pac-Man} & {\small{}71.7\%} & {\small{}505.5} & {\small{}704.9}\tabularnewline
\hline 
\textbf{\small{}Frostbite} & {\small{}62.6\%} & {\small{}114.6} & {\small{}183.1}\tabularnewline
\hline 
\textbf{\small{}Breakout} & {\small{}53.2\%} & {\small{}5.0} & {\small{}9.4}\tabularnewline
\hline 
\textbf{\small{}Road Runner} & {\small{}13.2\%} & {\small{}247.5} & {\small{}1870.0}\tabularnewline
\hline 
\textbf{\small{}Pong} & \multicolumn{3}{c|}{{\small{}no learning within tested 20K steps}}\tabularnewline
\hline 
\textbf{\small{}Zaxxon} & \multicolumn{3}{c|}{{\small{}no learning within tested 20K steps}}\tabularnewline
\hline 
\end{tabular}

\caption{Summary of experiment results}
\end{table}

For four games, the SAM-augmented agent showed better results (marked
in bold) than the raw pixel agent. Those games all belong to the ``easy
exploration -- high number of onscreen objects'' cluster. For eight
games, the raw pixel agent performed better than the SAM-augmented
agent. For two games, neither agent showed measurable learning progress.

\newpage{}

\section{Discussion}

\textbf{Contribution of this work to the domain of video game RL}:
\\\\
The current state of the art for video game RL uses raw pixels of
video games as input to the RL algorithm. The raw pixels from the
video games are preprocessed with a simple and computationally efficient
two-step preprocessing pipeline (downscale to 84x84 pixels; grayscale
image), before a five-layer convolutional neural network processes
it, which resembles the environment information for the RL agent (\cite{machado2018}).
The CleanRL Python framework, which is used as the basis for the experiments
in this research, has implemented this state of the art method (\cite{huang2022}).
This research investigated the impact of additional preprocessing
performed on the raw pixels before being processed by the state of
the art pipeline outlined above. Meta Research's SAM image segmentation
model (\cite{kirillov2023}) was added as an additional step into
the state of the art pipeline. Although the experiment results show
that only four out of twelve games tested could be improved with this
method, it has been demonstrated that such an approach does have an
impact on video game RL, and in some cases can actually improve video
game RL compared with the current state of the art pipeline. Atari
games have been clustered into easy and hard exploration difficulty
levels (\cite{bellemare2016}). This research added the criterion
``low or high number of onscreen objects. The best performance is
observed for the category ``easy exploration'', combined with the
criterion ``high number of on-screen objects''. For these four games
(Space Invaders, Chopper Command, Seaquest and Beam Rider), even with
only 20K training steps, already a distinct performance improvement
of the SAM-augmented RL agent over the raw pixels RL agent can be
observed. On the other hand, in general games with few and small objects
relevant for gameplay do not perform better, but worse with a SAM-augmented
RL agent. For the category ``hard exploration'' games, the SAM-augmented
RL agent performance was worse, with one potential exception of the
game ``Zaxxon''.

\textbf{Future Research}:
\\\\
The results show that adding image segmentation to video game reinforcement
learning, which segments the raw video game image into areas and objects,
shows potential to improve RL agent performance, but not for all types
of games, and not without further research on finetuning the use of
the SAM image segmentation model (or other similar models). For the
better-performing game agents, the performance lift comes with a heavy
computational price. While RL for video games without complex image
preprocessing can easily train agents up to tens or hundreds of millions
of training steps, the computational cost for image segmentation slows
down the training process immensely. Introducing a foundation model
like SAM into the RL pipeline increases the training time by a factor
of 500 or more. Partially this price can be addressed by better GPU
infrastructure and optimized ML model inference, but it will still
be much slower than traditional RL training without image preprocessing.

\textbf{Limitations}:
\\\\
Different SAM model inference hyperparameters lead to vastly different
perception of the game scene. Different sets of SAM parameters can
positively influence the RL agent performance, for example because
more objects relevant for gameplay are segmented, or (when using opacity)
not only the detected segments, but a combination of raw pixel and
segments is presented to the RL agent for perception. As an alternative
to using SAM, other image augmentation libraries, e.g. Scikit-Image,
CutLER or Detectron2, could be explored. The approach could be tried
on games of newer video game platforms than Atari 2600, e.g. Arcade
games (with MAMEToolkit) or 1990ies video game platforms like Sega
Mega Drive (with Stable-Retro). The graphics complexity of these platforms
is much higher than the ``almost mask-like simple graphics'' of
the Atari 2600 platform. Therefore the impact of identifying and masking
objects may be higher in more unstructured raw images like from these
newer video game platforms. 

\begin{center}
\begin{figure}[H]
\begin{centering}
\includegraphics[width=12cm]{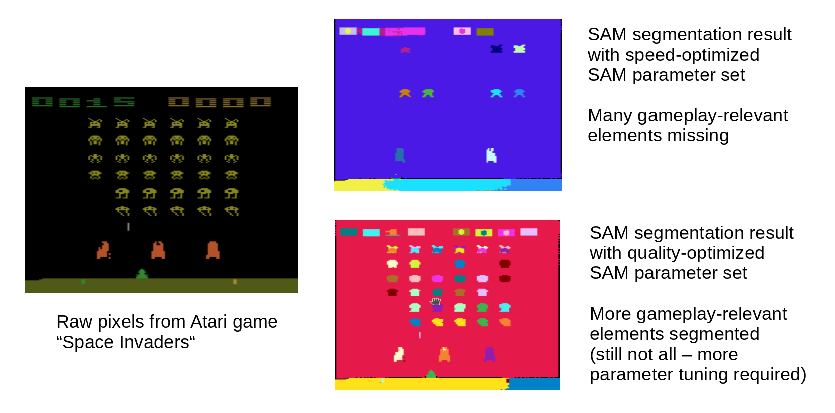}
\par\end{centering}
\caption{Impact of SAM hyperparameter tuning on image augmentation}
\end{figure}
\par\end{center}

\begin{center}
\begin{figure}[H]
\begin{centering}
\includegraphics[width=10cm]{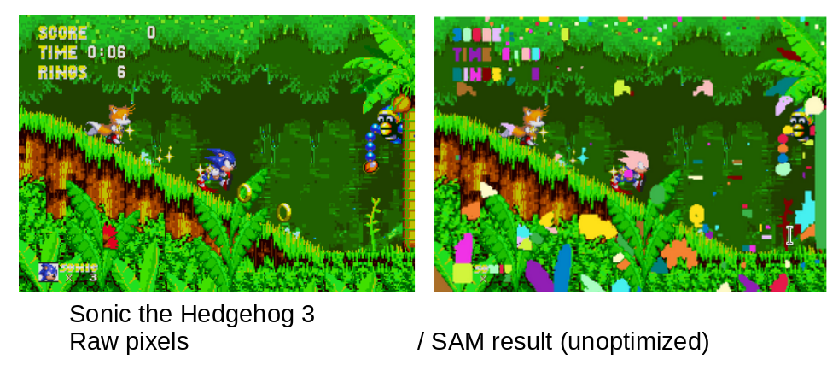}
\par\end{centering}
\caption{Potential usage of SAM on more complex video game scenes}
\end{figure}
\par\end{center}

\section{Conclusion}

Adding an ``image segmentation'' image preprocessing step to video
game RL does have an impact on RL performance. The experiment results
from adding Meta Research's SAM image segmentation model to the
state of the art video game RL pipeline show that it can be a positive
one under certain conditions such as a high number of gameplay objects
typically seen on screen during the game. However, using SAM comes
at a high computational price which slows down the RL training process
significantly. Therefore, it can be concluded that while this novel
approach shows potential, it is not recommended to follow it currently
as a new standard pipeline for video game RL without further research.
The focus of future research should be to find a good balance between
performance improvement through image augmentation and RL training
performance decline due to the additional computing requirements introduced
due to the additional preprocessing step. For this, fine-tuning the
hyperparameters of SAM model inference, and the usage of alternative
image augmentation algorithms, such as Scikit-Image, CutLER or Detectron2,
should be explored primarily. 

\newpage{}

\bibliography{var4arl}
\bibliographystyle{iclr2024_conference}

\appendix

\section{Code Repository}

The RL baseline installation documentation and the code implemented
for this paper are published at \url{https://github.com/c-a-schiller/var4arl}

\newpage{}

\section{Detailed results for each game}

For better readability of visualized learning curves, Tensorboard
is configured with smoothing=0.99, which is equivalent to visualizing
the median values of the individual steps training results. Learning
curves visualize learning of SAM-augmented agent (orange) and raw
pixel agent (violet) simultaneously.

\subsection{Result details - Breakout}

\begin{table}[H]
\begin{tabular}{|c|c|c|}
\hline 
\textbf{Breakout} & \textbf{RL agent performance} & \textbf{RL agent performance}\tabularnewline
\cline{1-1} 
 & \textbf{without SAM} & \textbf{with SAM}\tabularnewline
\hline 
\textbf{Training time} & 1,5 min & 19,5 hours\tabularnewline
\hline 
\textbf{End result} & 9,42 & 4,97\tabularnewline
\hline 
\textbf{Learning curves} & \multicolumn{2}{c|}{\includegraphics[width=11.31cm]{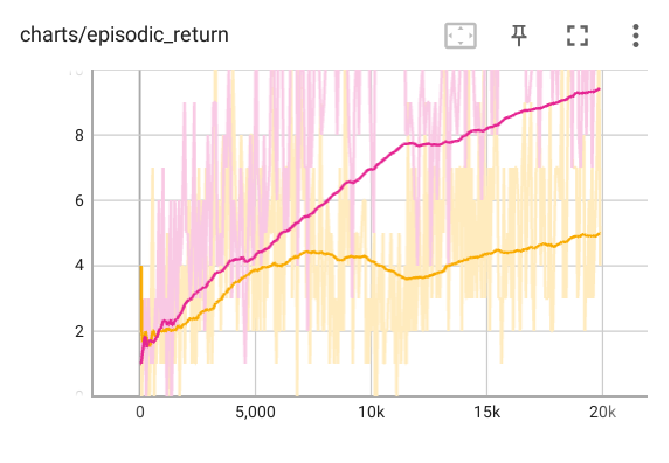}}\tabularnewline
\hline 
\textbf{Example frame} & \includegraphics[width=5cm,height=3.75cm]{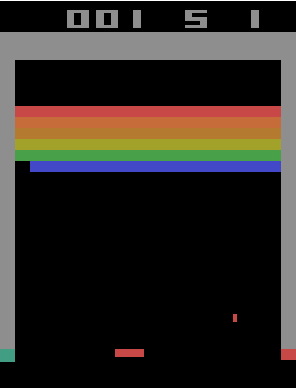} & \includegraphics[width=5cm,height=3.75cm]{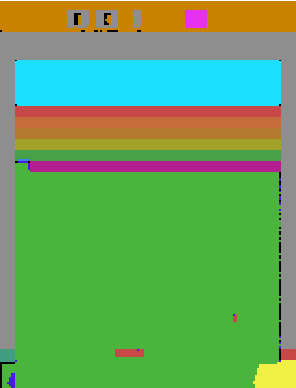}\tabularnewline
\hline 
\end{tabular}

\textbf{Interpretation:} SAM-augmented agent performs worse than the
raw image processing agent. The SAM-segmented game image is, apart
from the colors, nearly indistinguishable from the original image,
which, due to its abstractness, is already almost ``mask-like''
itself. But in some areas SAM adds noise, e.g. around the score counters,
on the right side and in the bottom right corner. This additional
noise could explain why the SAM-augmented agent learns less well than
the raw image processing agent.

\caption{Result details - Breakout}
\end{table}

\subsection{Result details - Pong}

\begin{table}[H]
\begin{tabular}{|c|c|c|}
\hline 
\textbf{Pong} & \textbf{RL agent performance} & \textbf{RL agent performance}\tabularnewline
\cline{1-1} 
 & \textbf{without SAM} & \textbf{with SAM}\tabularnewline
\hline 
\textbf{Training time} & 1,2 min & 15,86 hours\tabularnewline
\hline 
\textbf{End result} & -21 & -20,93\tabularnewline
\hline 
\textbf{Learning curves} & \multicolumn{2}{c|}{\includegraphics[width=11.31cm]{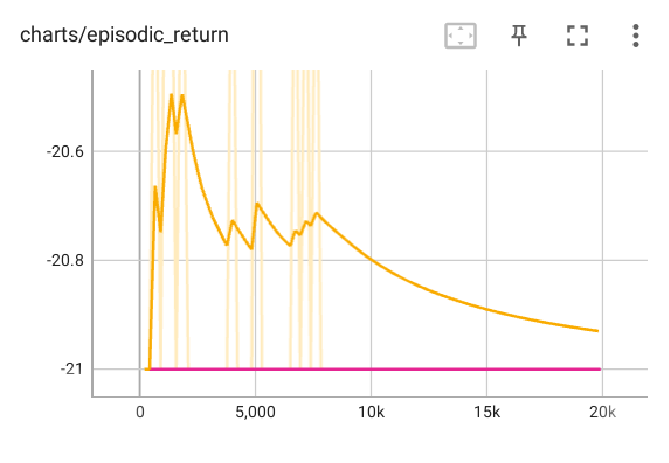}}\tabularnewline
\hline 
\textbf{Example frame} & \includegraphics[width=5cm,height=3.75cm]{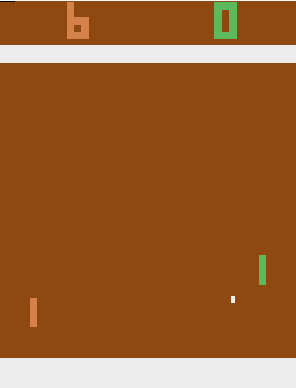} & \includegraphics[width=5cm,height=3.75cm]{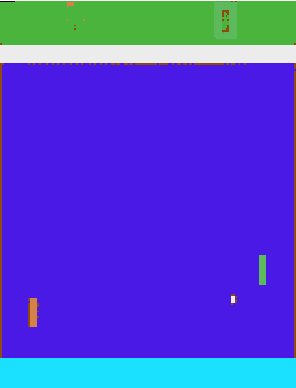}\tabularnewline
\hline 
\end{tabular}

\textbf{Interpretation:} SAM-augmented agent performs marginally better
than raw image processing agent. The SAM-segmented game image is,
apart from the colors, nearly indistinguishable from the original
image, which, due to its abstractness, is already almost ``mask-like''
itself. In certain areas SAM adds noise, e.g. around the score counters.
However, the performance improvement of SAM vs. non-SAM agent is marginal.
With the fixed set of hyperparameters, neither agent learned to play
the game at all.

\caption{Result details - Pong}
\end{table}

\subsection{Result details - Space Invaders}

\begin{table}[H]
\begin{tabular}{|c|c|c|}
\hline 
\textbf{Space Invaders} & \textbf{RL agent performance} & \textbf{RL agent performance}\tabularnewline
\cline{1-1} 
 & \textbf{without SAM} & \textbf{with SAM}\tabularnewline
\hline 
\textbf{Training time} & 1,3 min & 16,9 hours\tabularnewline
\hline 
\textbf{End result} & 223,3 & 226,2\tabularnewline
\hline 
\textbf{Learning curves} & \multicolumn{2}{c|}{\includegraphics[width=11.31cm]{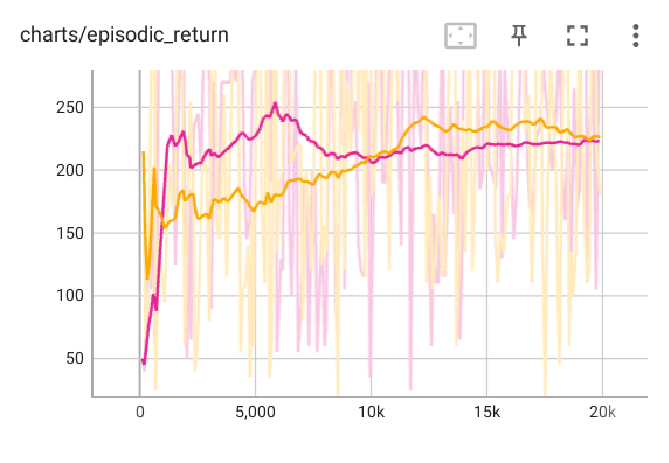}}\tabularnewline
\hline 
\textbf{Example frame} & \includegraphics[width=5cm,height=3.75cm]{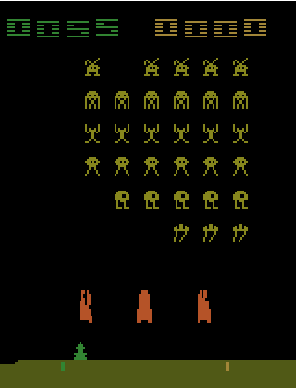} & \includegraphics[width=5cm,height=3.75cm]{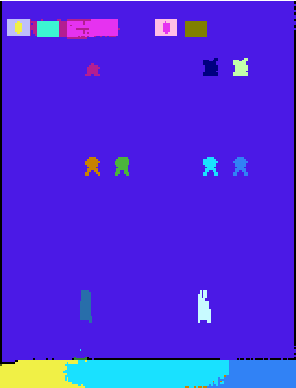}\tabularnewline
\hline 
\end{tabular}

\textbf{Interpretation:} SAM-augmented agent performs slightly better
than the raw image processing agent. The SAM-transformed image shows
that with the SAM option set chosen due to inference time reasons,
which reduces the image segmentation quality of the model, the SAM
model is not able to detect all objects relevant for gameplay. SAM
model options would need to be adapted, at the cost of inference performance.

\caption{Result details - Space Invaders}
\end{table}

\subsection{Result details - Chopper Command}

\begin{table}[H]
\begin{tabular}{|c|c|c|}
\hline 
\textbf{Chopper Command} & \textbf{RL agent performance} & \textbf{RL agent performance}\tabularnewline
\cline{1-1} 
 & \textbf{without SAM} & \textbf{with SAM}\tabularnewline
\hline 
\textbf{Training time} & 1,4 min & 16,5 hours\tabularnewline
\hline 
\textbf{End result} & 889,6 & 932,2\tabularnewline
\hline 
\textbf{Learning curves} & \multicolumn{2}{c|}{\includegraphics[width=11.31cm]{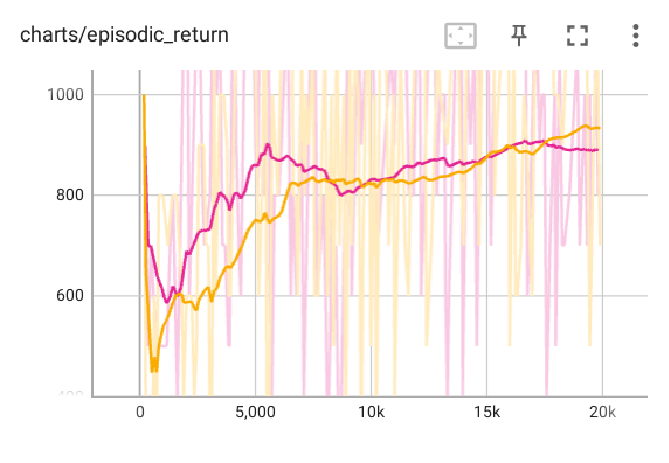}}\tabularnewline
\hline 
\textbf{Example frame} & \includegraphics[width=5cm,height=3.75cm]{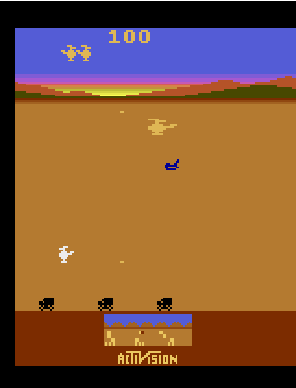} & \includegraphics[width=5cm,height=3.75cm]{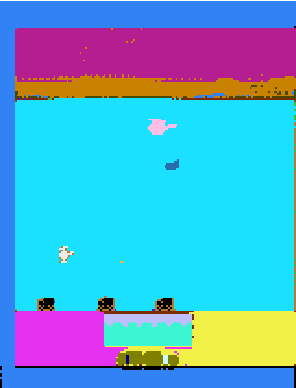}\tabularnewline
\hline 
\end{tabular}

\textbf{Interpretation:} SAM-augmented agent performs slightly better
than the raw image processing agent. The SAM-transformed image shows
that most gameplay-relevant elements are identified, including the
small bullets.

\caption{Result details - Chopper Command}
\end{table}

\subsection{Result details - Kung Fu Master}

\begin{table}[H]
\begin{tabular}{|c|c|c|}
\hline 
\textbf{Kung Fu Master} & \textbf{RL agent performance} & \textbf{RL agent performance}\tabularnewline
\cline{1-1} 
 & \textbf{without SAM} & \textbf{with SAM}\tabularnewline
\hline 
\textbf{Training time} & 1,3 min & 15,0 hours\tabularnewline
\hline 
\textbf{End result} & 78,4 & 73,2\tabularnewline
\hline 
\textbf{Learning curves} & \multicolumn{2}{c|}{\includegraphics[width=11.31cm]{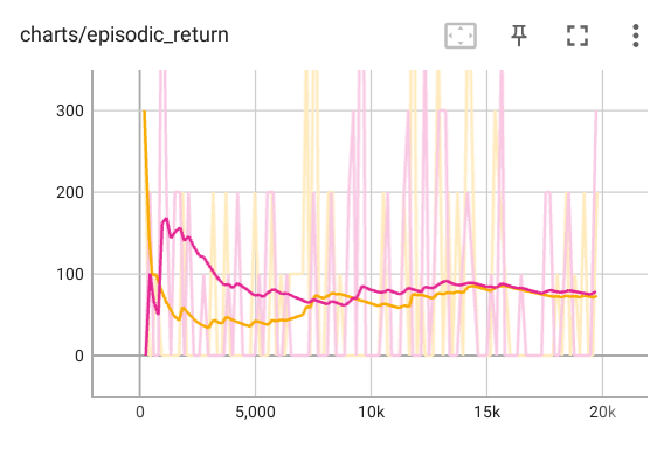}}\tabularnewline
\hline 
\textbf{Example frame} & \includegraphics[width=5cm,height=3.75cm]{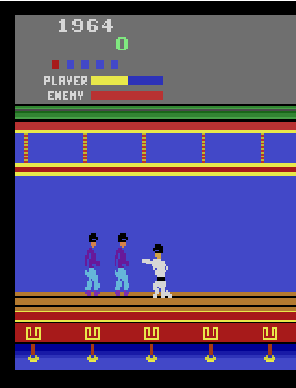} & \includegraphics[width=5cm,height=3.75cm]{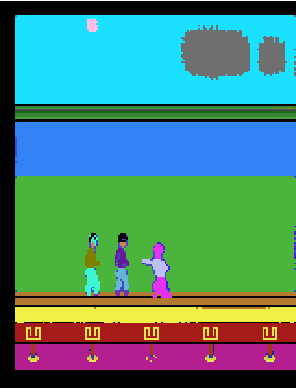}\tabularnewline
\hline 
\end{tabular}

\textbf{Interpretation:} SAM-augmented agent performs slightly worse
than the raw image processing agent. The SAM-transformed image shows
that most gameplay-relevant elements are identified. However some
unexplained noise is present, like the dark ``clouds'' top-right,
even though there is no apparent reason for this in the original image.

\caption{Result details - Kung Fu Master}
\end{table}

\subsection{Result details - Road Runner}

\begin{table}[H]
\begin{tabular}{|c|c|c|}
\hline 
\textbf{Road Runner} & \textbf{RL agent performance} & \textbf{RL agent performance}\tabularnewline
\cline{1-1} 
 & \textbf{without SAM} & \textbf{with SAM}\tabularnewline
\hline 
\textbf{Training time} & 1,54 min & 18,2 hours\tabularnewline
\hline 
\textbf{End result} & 1870,0 & 247,5\tabularnewline
\hline 
\textbf{Learning curves} & \multicolumn{2}{c|}{\includegraphics[width=11.31cm]{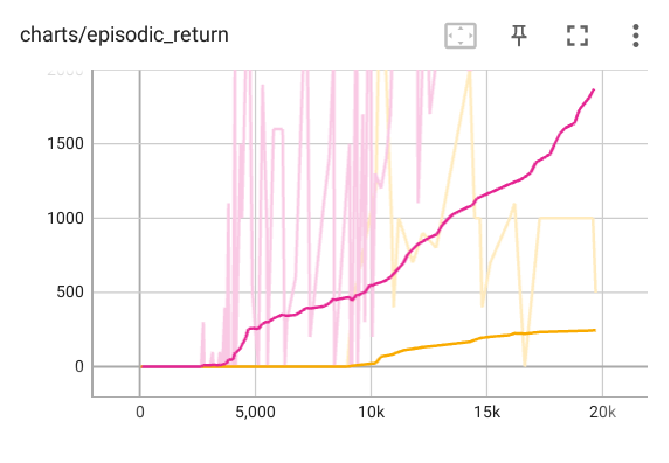}}\tabularnewline
\hline 
\textbf{Example frame} & \includegraphics[width=5cm,height=3.75cm]{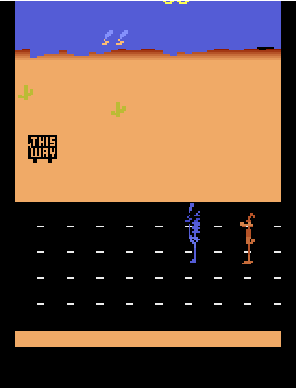} & \includegraphics[width=5cm,height=3.75cm]{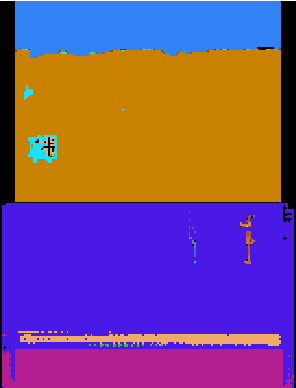}\tabularnewline
\hline 
\end{tabular}

\textbf{Interpretation:} SAM-augmented agent performs much worse than
the raw image processing agent. The SAM-transformed image shows that
gameplay-relevant elements are identified not well, e.g. the player
character (blue) is almost invisible in the SAM-transformed image.

\caption{Result details - Road Runner}
\end{table}

\subsection{Result details - Seaquest}

\begin{table}[H]
\begin{tabular}{|c|c|c|}
\hline 
\textbf{Seaquest} & \textbf{RL agent performance} & \textbf{RL agent performance}\tabularnewline
\cline{1-1} 
 & \textbf{without SAM} & \textbf{with SAM}\tabularnewline
\hline 
\textbf{Training time} & 1,3 min & 16,9 hours\tabularnewline
\hline 
\textbf{End result} & 218,1 & 230,4\tabularnewline
\hline 
\textbf{Learning curves} & \multicolumn{2}{c|}{\includegraphics[width=11.31cm]{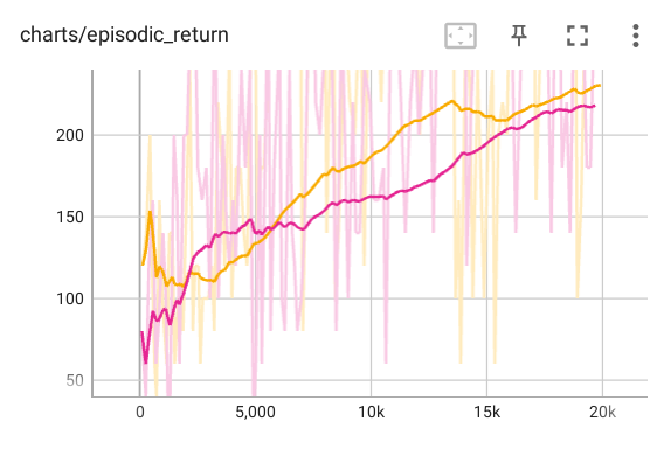}}\tabularnewline
\hline 
\textbf{Example frame} & \includegraphics[width=5cm,height=3.75cm]{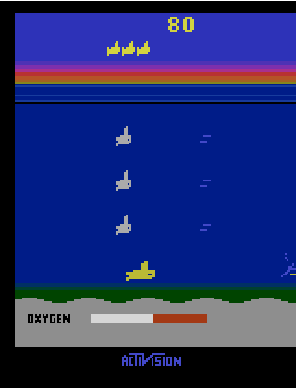} & \includegraphics[width=5cm,height=3.75cm]{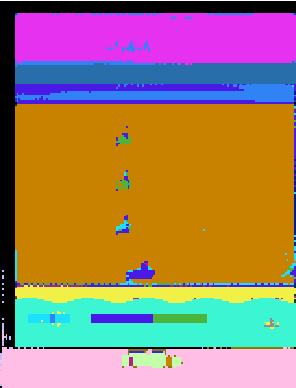}\tabularnewline
\hline 
\end{tabular}

\textbf{Interpretation:} SAM-augmented agent performs better than
the raw image processing agent, even though some gameplay-relevant
objects are not segmented, like the light blue characters in the middle
of the screen. 

\caption{Result details - Seaquest}
\end{table}

\subsection{Result details - Beam Rider}

\begin{table}[H]
\begin{tabular}{|c|c|c|}
\hline 
\textbf{Beam Rider} & \textbf{RL agent performance} & \textbf{RL agent performance}\tabularnewline
\cline{1-1} 
 & \textbf{without SAM} & \textbf{with SAM}\tabularnewline
\hline 
\textbf{Training time} & 1,3 min & 15,5 hours\tabularnewline
\hline 
\textbf{End result} & 390,3 & 505,1\tabularnewline
\hline 
\textbf{Learning curves} & \multicolumn{2}{c|}{\includegraphics[width=11.31cm]{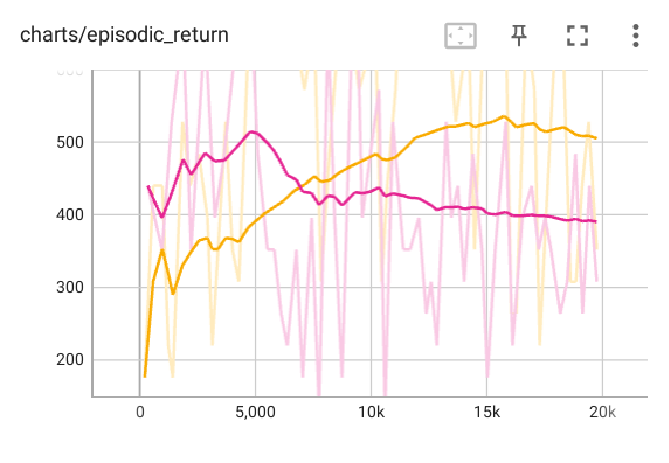}}\tabularnewline
\hline 
\textbf{Example frame} & \includegraphics[width=5cm,height=3.75cm]{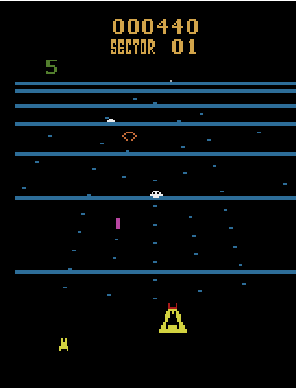} & \includegraphics[width=5cm,height=3.75cm]{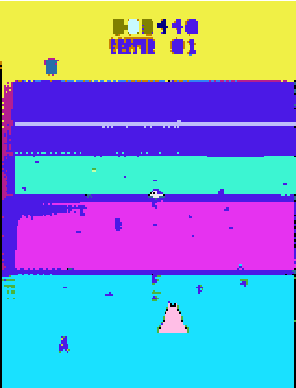}\tabularnewline
\hline 
\end{tabular}

\textbf{Interpretation:} SAM-augmented agent performs better than
the raw image processing agent, even though some gameplay-relevant
objects are not segmented and parts of the background are segmented
instead.

\caption{Result details - Beam Rider}
\end{table}

\subsection{Result details - Ms. Pac-Man}

\begin{table}[H]
\begin{tabular}{|c|c|c|}
\hline 
\textbf{Ms. Pac-Man} & \textbf{RL agent performance} & \textbf{RL agent performance}\tabularnewline
\cline{1-1} 
 & \textbf{without SAM} & \textbf{with SAM}\tabularnewline
\hline 
\textbf{Training time} & 1,3 min & 15,4 hours\tabularnewline
\hline 
\textbf{End result} & 704,9 & 505,5\tabularnewline
\hline 
\textbf{Learning curves} & \multicolumn{2}{c|}{\includegraphics[width=11.31cm]{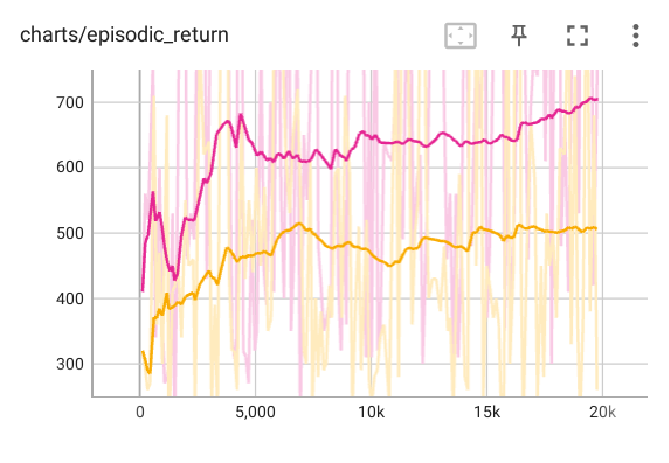}}\tabularnewline
\hline 
\textbf{Example frame} & \includegraphics[width=5cm,height=3.75cm]{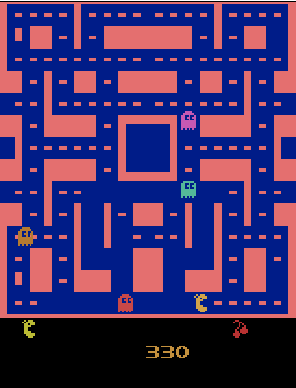} & \includegraphics[width=5cm,height=3.75cm]{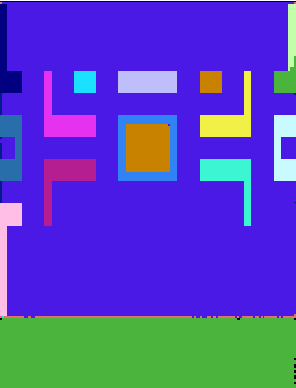}\tabularnewline
\hline 
\end{tabular}

\textbf{Interpretation:} SAM-augmented agent performs worse than the
raw image processing agent. The SAM image segmentation totally fails
to pick up the gameplay-relevant elements.

\caption{Result details - Ms. Pac-Man}
\end{table}

\subsection{Result details - Q{*}Bert}

\begin{table}[H]
\begin{tabular}{|c|c|c|}
\hline 
\textbf{Q{*}Bert} & \textbf{RL agent performance} & \textbf{RL agent performance}\tabularnewline
\cline{1-1} 
 & \textbf{without SAM} & \textbf{with SAM}\tabularnewline
\hline 
\textbf{Training time} & 1,4 min & 17,6 hours\tabularnewline
\hline 
\textbf{End result} & 456,6 & 375,8\tabularnewline
\hline 
\textbf{Learning curves} & \multicolumn{2}{c|}{\includegraphics[width=11.31cm]{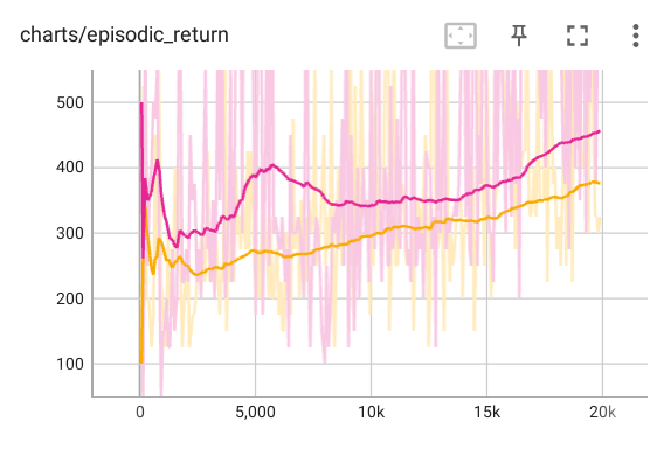}}\tabularnewline
\hline 
\textbf{Example frame} & \includegraphics[width=5cm,height=3.75cm]{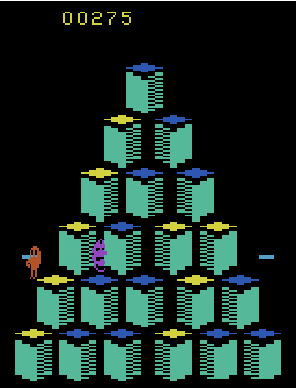} & \includegraphics[width=5cm,height=3.75cm]{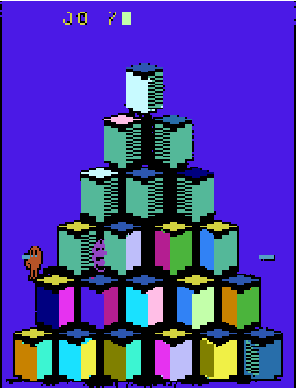}\tabularnewline
\hline 
\end{tabular}

\textbf{Interpretation:} SAM-augmented agent performs worse than the
raw image processing agent. The SAM image segmentation picks up background
elements instead of gameplay-relevant character elements.

\caption{Result details - Q{*}Bert}
\end{table}

\subsection{Result details - Frostbite}

\begin{table}[H]
\begin{tabular}{|c|c|c|}
\hline 
\textbf{Frostbite} & \textbf{RL agent performance} & \textbf{RL agent performance}\tabularnewline
\cline{1-1} 
 & \textbf{without SAM} & \textbf{with SAM}\tabularnewline
\hline 
\textbf{Training time} & 1,5 min & 18,2 hours\tabularnewline
\hline 
\textbf{End result} & 183,1 & 114,6\tabularnewline
\hline 
\textbf{Learning curves} & \multicolumn{2}{c|}{\includegraphics[width=11.31cm]{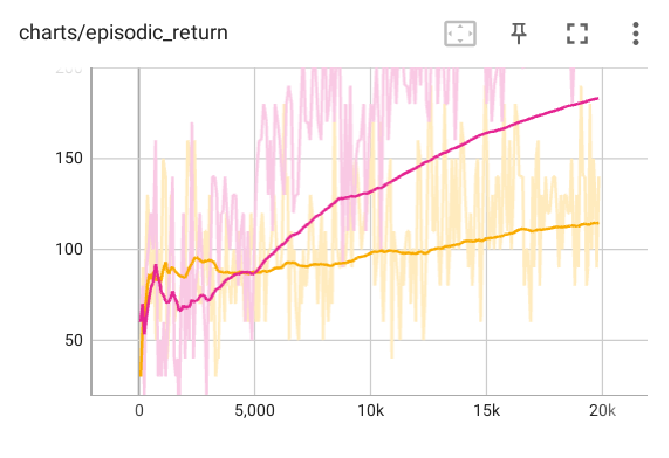}}\tabularnewline
\hline 
\textbf{Example frame} & \includegraphics[width=5cm,height=3.75cm]{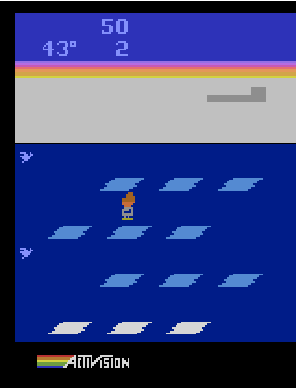} & \includegraphics[width=5cm,height=3.75cm]{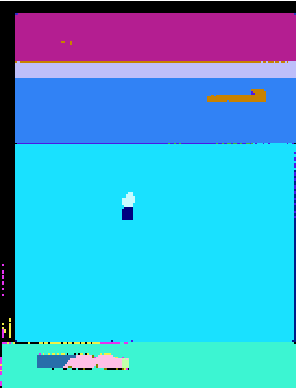}\tabularnewline
\hline 
\end{tabular}

\textbf{Interpretation:} SAM-augmented agent performs worse than the
raw image processing agent. The SAM image segmentation has trouble
picking up gameplay-relevant elements, such as the platforms the player
has to jump upon. The whole playing area is segmented as blue.

\caption{Result details - Frostbite}
\end{table}

\subsection{Result details - Zaxxon}

\begin{table}[H]
\begin{tabular}{|c|c|c|}
\hline 
\textbf{Zaxxon} & \textbf{RL agent performance} & \textbf{RL agent performance}\tabularnewline
\cline{1-1} 
 & \textbf{without SAM} & \textbf{with SAM}\tabularnewline
\hline 
\textbf{Training time} & 1,3 min & 15,9 hours\tabularnewline
\hline 
\textbf{End result} & 2,9 & 31,4\tabularnewline
\hline 
\textbf{Learning curves} & \multicolumn{2}{c|}{\includegraphics[width=11.31cm]{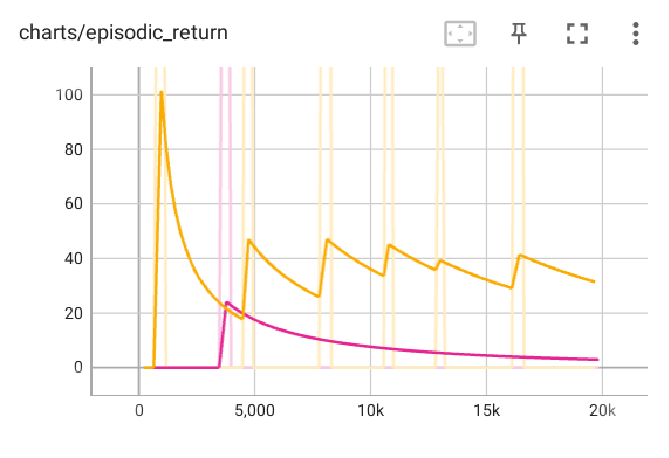}}\tabularnewline
\hline 
\textbf{Example frame} & \includegraphics[width=5cm,height=3.75cm]{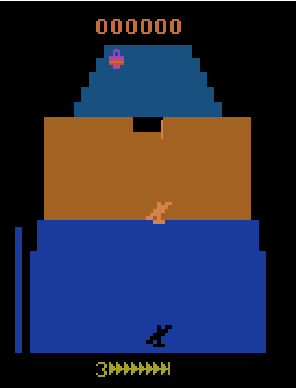} & \includegraphics[width=5cm,height=3.75cm]{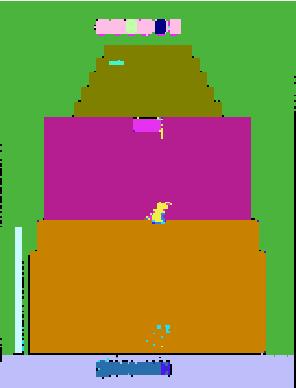}\tabularnewline
\hline 
\end{tabular}

\textbf{Interpretation:} SAM-augmented agent performs better than
the raw image processing agent, though neither agent learned to play
the game in the 20K steps performed in this research. The SAM image
segmentation shows that gameplay-relevant character elements are segmented,
so it could be possible that with more steps performed the SAM-augmented
agent may perform better than the raw image processing agent.

\caption{Result details - Zaxxon}
\end{table}

\end{document}